\documentclass[10pt,twocolumn,letterpaper]{article}

\usepackage{cvpr}
\usepackage{times}
\usepackage{epsfig}
\usepackage{graphicx}
\usepackage{amsmath}
\usepackage{amssymb}

\usepackage{subfigure}
\usepackage{float}
\usepackage{rotating}

\usepackage[pagebackref=true,breaklinks=true,letterpaper=true,colorlinks,bookmarks=false]{hyperref}

 \cvprfinalcopy 

\ifcvprfinal\fi

\begin{document}
\title{\ Deep Inception Generative Network for Cognitive Image Inpainting}

\author{Qingguo Xiao,Guangyao Li,Qiaochuan Chen
\\
Tongji University, Shanghai, China
\\
lgy423@126.com
}


\maketitle
\begin{figure*}[!htbp]
\begin{center}
\subfigure{\includegraphics[width=0.22\linewidth]{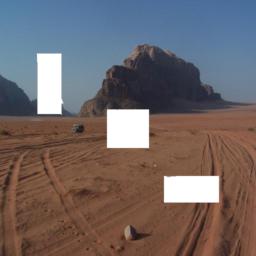}}
\subfigure{\includegraphics[width=0.22\linewidth]{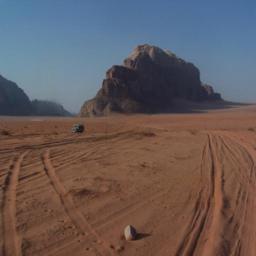}}
\subfigure{\includegraphics[width=0.22\linewidth]{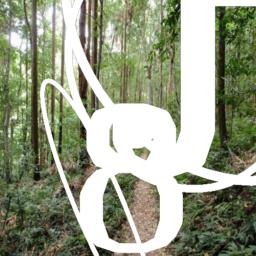}}
\subfigure{\includegraphics[width=0.22\linewidth]{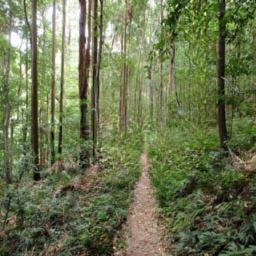}}
\vspace{-0.3cm}
\subfigure{\includegraphics[width=0.22\linewidth]{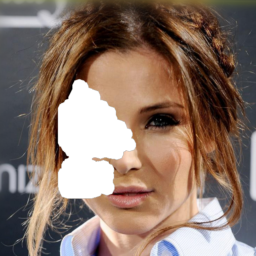}}
\subfigure{\includegraphics[width=0.22\linewidth]{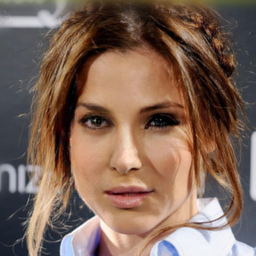}}
\subfigure{\includegraphics[width=0.22\linewidth]{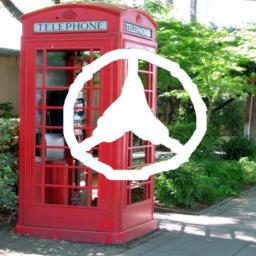}}
\subfigure{\includegraphics[width=0.22\linewidth]{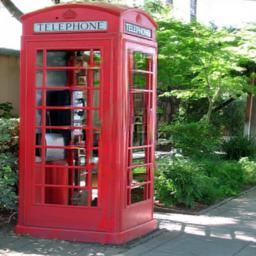}}
\end{center}
   \caption{ Image completion results by our deep inception generative network. There are several different masks including rectangles , irregular regions, and customized shapes which are draw by hand on canvas of PhotoShop.}
\label{Figure 1}
\end{figure*}

\begin{abstract}
Recent advances in deep learning have shown exciting promise in filling large holes and lead to another orientation for image inpainting. However, existing learning-based methods often create artifacts and fallacious textures because of insufficient cognition understanding. Previous generative networks are limited with single receptive type and give up pooling in consideration of detail sharpness. 修改不该也行  Human cognition is constant regardless of the target attribute. As multiple receptive fields improve the ability of abstract image characterization  and pooling can keep feature invariant, specifically, deep inception learning is adopted to promote high-level feature representation and enhance model learning capacity for local patches. Moreover, approaches for generating diverse mask images are introduced and a random mask dataset is created. We benchmark our methods on ImageNet, Places2 dataset, and CelebA-HQ. Experiments for regular, irregular, and custom regions completion are all performed and free-style image inpainting is also presented. Quantitative comparisons with previous state-of-the-art methods show that ours obtain much more natural image completions.

\end{abstract} 
\section{Introduction}
Image inpainting is a technique that aims to restore the damaged images or fill in the missing parts of images with visually plausible contents \cite{barnes2009patchmatch,criminisi2004region}. It allows to remove distracting objects or retouch undesired regions in photos \cite{darabi2012image,he2014image}. It can be extended to other image generative tasks such as super-resolution \cite{huang2015single, shi2016real}. And  it is  an important step in many graphics algorithms, e.g.,  generating a clean background plate or reshuffling image contents\cite{huang2014image}. Due to the inherent ambiguity and complexity of natural images, image completion is a challenging problem. 

Recent years, Deep learning  has been applied to this troublesome task attributed to the outstanding performance. Some image generative networks are designed for hole-filling based on Convolutional Neural Networks (CNN). \cite{pathak2016context} trained an encoder-decoder CNN (Context Encoder). It is a pioneering model. \cite{iizuka2017globally} designed global and local context discriminators to distinguish real images from completed ones using Generative Adversarial Network (GAN) framework. However, a simple fact currently is that compared with various types of networks in the domain of image classification and recognition where the models vary from squeezing AlexNet for FPGA and embedded deployment \cite{iandola2016squeezenet} to very large scale networks with several hundred layers \cite{szegedy2017inception}, networks for image inpainting are few. Moreover, some models even come from the domain of image segmentation, super resolution, and image style transfer. \cite{yeh2017semantic} used the DCGAN model architecture from \cite{radford2015unsupervised}.  The generative model used in \cite{yan2018shift} and \cite{liu2018image} is based on that in \cite{isola2017image} which is for image-to-image translation task. 

In addition, there are two main problems which current image completion faces to. First, these methods often create boundary artifacts, distorted structures and blurry textures inconsistent with surrounding areas because of insufficient cognitive understanding and ineffectiveness of convolutional neural networks in modeling long-term correlations between contextual information and the hole regions \cite{pathak2016context,yu2018generative,iizuka2017globally}. For example, the target is a dog, but the completed result does not follow the vision cognition. It does not look like a dog visually and much details are artifacts. The filter used by traditional CNNs is a generalized linear model (GLM). Therefore, it is implicitly assumed that the features are linearly separable for extraction, but the actual case is often difficult to be linearly separable. Furthermore,  most generative networks give up pooling and are limited with $3\times3$ convolutional kernels. This is obviously not possible to fully utilize its learning ability and cognitive understanding due to using only a single type of receptive fields.
Given a dog in an image,  from our human vision cognition, it is always a dog no matter where the target is, big or small it is, and rotated or not it is. Vision cognition keeps invariable. Specifically,  deep inception learning is adopted to utilize more complex structures to abstract the data within diverse receptive fields and explore enough cognitive understanding.  Micro neural networks are build within a layer and the inpainting network can be constructed by stacking multiple of these layers for efficient high-level feature extraction by means of the ability of nonlinearity of inception layer.

Another problem is that previous deep learning approaches focused on rectangular regions located around the center of the image\cite{pathak2016context,yang2017high,yan2018shift}. Several works for random inpainting are still limited to regular shape masks and rectangle region is the most used form.  \cite{liu2018image} created an irregular mask dataset for train inpainting networks using the research work in \cite{sundaram2010dense}. However it is insufficient for arbitrary completion and free-style inpainting. It is necessary to include regular, special, and any other style shapes apart from irregular masks. To address this limitation, methods to create diver masks for model robustness of arbitrary completion are introduced in this paper. Combined with the above constructed network,  free-style inpainting is finally realized.

Our contributions  are summarized as follows. First, a novel generative network  architecture using inception modules is proposed to enhance the abstraction ability of feature.
The constructed network  significantly improves completion results. To the best of our knowledge, we are the first to adopt inception learning for image inpainting. Moreover, approaches for generating diverse masks are provided and a relevant mask dataset is created. A variety of comparative experiments are performed on benchmark datasets. High-quality inpainting results are achieved and results demonstrate that our model is robust for arbitrary completion including on regular, irregular, and custom masks as shown in Figure \ref{Figure 1}.

\section{Related Work}
A variety of  approaches have been proposed for the image completion task. There are two mainstreams.
\subsection{Examplar-based Inpainting}
One category of traditional image completion methods are exemplar-based. They have been the main method for a long time. This task is a patch-based optimization problem with an energy function \cite{criminisi2004region,wexler2007space,barnes2009patchmatch,darabi2012image,he2014image,huang2014image}. Matching-based methods explicitly match the patches in the unknown region with the patches in the known region, and copy the known content to complete the unknown region. The obvious limitation  is that the synthesized texture only comes from the input image. This is a problem when a convincing completion requires textures that are not found in the input image. It is the same for diffusion-based methods which solve Partial Differential Equations (PDE) \cite{bertalmio2000image,levin2003learning} and propagate colors into the missing regions.

\subsection{Learning-based Inpainting}  
More recently, deep neural networks are introduced for texture synthesis and image stylization \cite{isola2017image,zhu2017unpaired}. Deep learning methods provide encouraging results in filling large target regions \cite{pathak2016context,yang2017high,yeh2017semantic,liu2017semantically,iizuka2017globally}. 
In particular, \cite{pathak2016context} train an Encoder-Decoder CNN (Context Encoder) with combined l2 and adversarial loss to directly predict plausible image structures. It has been a classic work for image inpainting using CNN. Dilated convolutions are adopted in \cite{iizuka2017globally} for increasing receptive fields of output neurons in inpainting network to replace channel-wise fully connected layer adopted in Context Encoder. In addition, global and local discriminators as adversarial losses based on GAN framework are introduced and  Poisson Blending is applied as a post-process. Based on the architecture of the deep convolutional generative
adversarial networks (DCGAN) from \cite{radford2015unsupervised}, semantic image inpainting is introduced in \cite{yeh2017semantic} to fill in the missing part conditioned on the known region for images from a specific semantic class. However, existing inpainting networks do not possess nonlinearity for a convolutional layer and these approaches are mainly trained on rectangular masks. 
\begin{figure}[!htbp]
\begin{center}
    \includegraphics[width=0.4\linewidth,height=3cm]{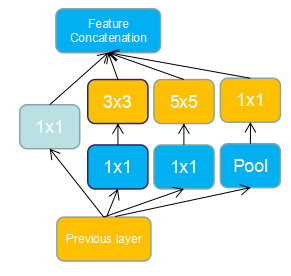}
\end{center}
\vspace{-0.3cm}
   \caption{One inception form.}
\label{Figure 2}
\end{figure}

Mask samples are essential for training in image inpainting task. The mainstream practice is to generate a random-sized broken area at a random position of the image. While the limitation of previous works is that the damaged area is mostly regular, and the rectangular block is most adopted \cite{pathak2016context,yang2017high,yan2018shift,iizuka2017globally}. A recent research in  \cite{sundaram2010dense} can generate masks of diverse shapes and stripes based on the results of mask image estimation between two consecutive frames of video. Combined with this research, partial convolution where the convolution is masked and re-normalized to utilize valid pixels is first proposed in \cite{liu2018image}  to better handle irregular masks.
\begin{figure*}[!htbp]
\begin{center}
    \includegraphics[width=0.7\linewidth,height=6cm]{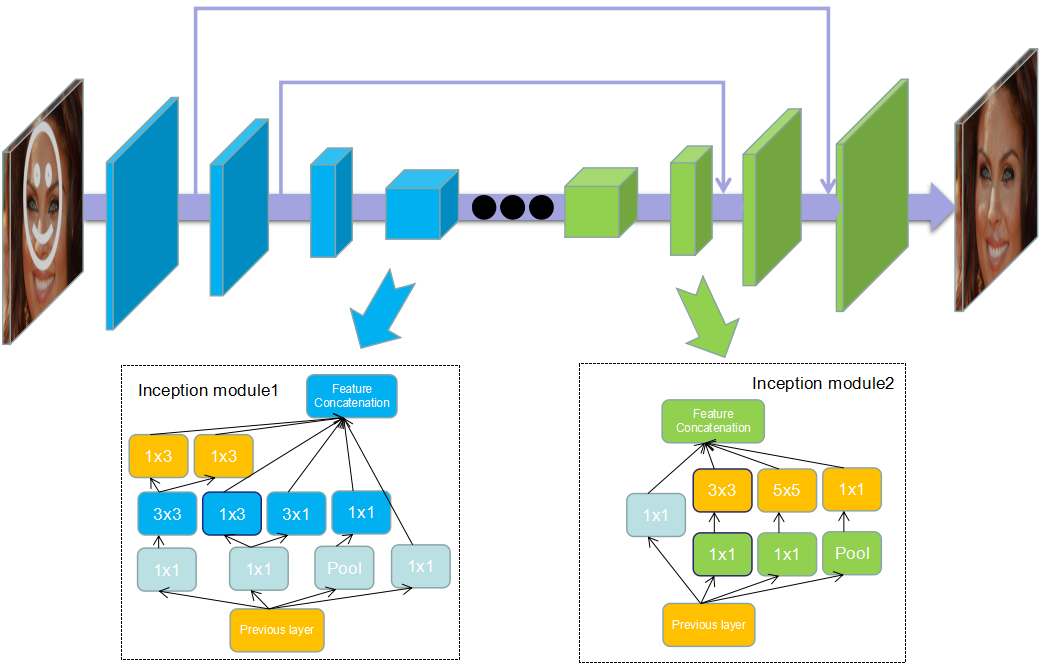}
\end{center}
\vspace{-0.3cm}
   \caption{Architecture of the image generative network. We follow the typical form of Encoder-Decoder. U-net is adopted to expand the learned features in encoder into the decoder. Our convolutional layer unit is an inception module and two kinds of inception modules are illustrated here. 
   Fed with an image with holes, the generator gives a corresponding completed one.}
\label{Figure 3}
\end{figure*}
\subsection{Network in Network}  
As stated above, the filter used by traditional CNNs is a generalized linear model. A solution is stacking convolution filters to generate higher-level feature representations to deal with practical problems and deeper models improve the abstract ability. Generally, the deeper the network is, the stronger the nonlinearity is. In general, the most direct way to improve network performance is increasing the depth and width of the network. But this way has the following problems: it brings too many parameters, and if the training data set is limited, it is easy to produce over-fitting; The larger the network is, the greater the computational complexity is; deeper network is prone to leads to gradient diffusion and the model is difficult to optimize. \cite{bengio2013representation} thinks that special design can be done in the convolutional layer so that the network can extract better features apart from stacking network convolutional layers. \cite{lin2013network} first introduces the idea of 'Network in Network' (NIN). A micro neural network which is a potent function approximator inside a convolutional layer is instantiated.   

\begin{figure}  
  \centering
    \includegraphics[width=0.7\linewidth]{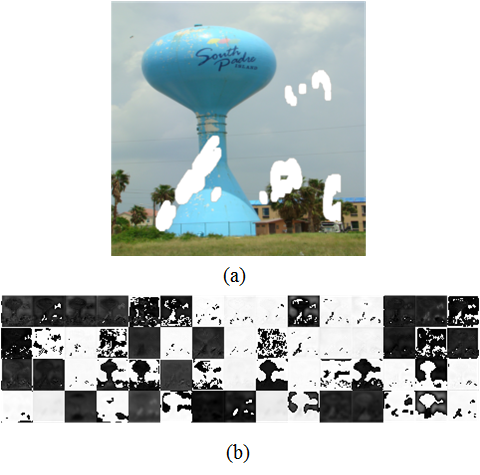} 
\caption{Visualization of the learned feature maps of different filters in the inception from the bottom fourth layer in our model. The images from top to bottom are respectively visuals of the first 15 channels' outputs using 1x1, 5x5, 3x3 convolutional kernel and pooling filter for the inception of the bottom fourth layer in the decoder.}
  \label{Figure 4} 
\end{figure}
Afterwards, inception modules are designed based on the framework of 'Network in Network'\cite{szegedy2015going, ioffe2015batch,szegedy2016rethinking}. The main idea of inception  is that an optimized local sparse structure in a convolutional neural network can be approximated and covered by a series of readily available dense substructures. Inception maintains the sparseness of the network and takes advantage of the high computational performance of dense matrices at the same time. It helps network to be deeper and wider without increasing the computation dramatically. Performance is improved significantly with the ability to extract more features under the same amount of computation. Combined with the design of the Residual Block Network (ResNet), which helps to mitigate the difficulty of training very deep networks,  a novel inception framework of  Inception-ResNet which utilizes the advantages of both Inception and ResNet is proposed in \cite{szegedy2017inception}. 
\section{Approach}
Our approach is based on deep inception learning, partial convolution, and random masks generating.
\subsection{Deep Inception Learning}   
Inspired by the above mentioned  'Network in Network' and 'Inception', deep inception learning is adopted for building our completion network. Inception is a micro network inside a layer and the NIN structure can  utilize strong nonlinearity than vanilla convolutions in the same receptive field.

There are several different types of kernels in one inception unit. Generally, a small filter, a medium-sized filter, a large filter, and a pooling filter are consisted. It increases the width of the network. On the other hand, it also increases the adaptability of the network for multi-scale processing. The network in the convolutional layer is able to extract information from each detail of the input, and large filter can also cover larger regions of the receiving layers. Pooling is performed to reduce the size of the space and  over-fitting. The topology of inception analyzes the relevant statistics of the upper layer and aggregates them into a highly related unit group. All the results are concatenated into a very deep feature map and the stitching means the fusion of different features.
There are many different inception modules and different combinations of filters in inceptions. Generally, there are a lot of stacks in the structure of inception and they express a dense and compressed form of information. To express it more sparsely and only aggregate the information in large quantities, $1\times1$ convolution is processed before general convolutions such as $3\times3$, $5\times5$. This helps reduce feature map thickness. For example, the output of the previous layer is $100\times100\times128$, the output is $100\times100\times256$ after a $5\times5$ convolutional layer with $256$ channels (stride=1, pad=2), and the final convolution layer parameters will be $128\times5\times5\times256 = 819200$. If the output of the previous layer passes through a $1\times1$ convolutional layer with $32$ channels first and then trace the $5\times5$ convolutional layer with $256$ channels, the output is still $100\times100\times256$, but the convolution parameter has been reduced to $128\times1\times1\times32 + 32\times5\times5\times256 = 204800$ which is a reduction of $4$ times. At the same time, more features can be extracted by superimposing more convolutions in the same receptive field.
\begin{figure}[!htbp]
\begin{center}
    \includegraphics[width=1.0\linewidth]{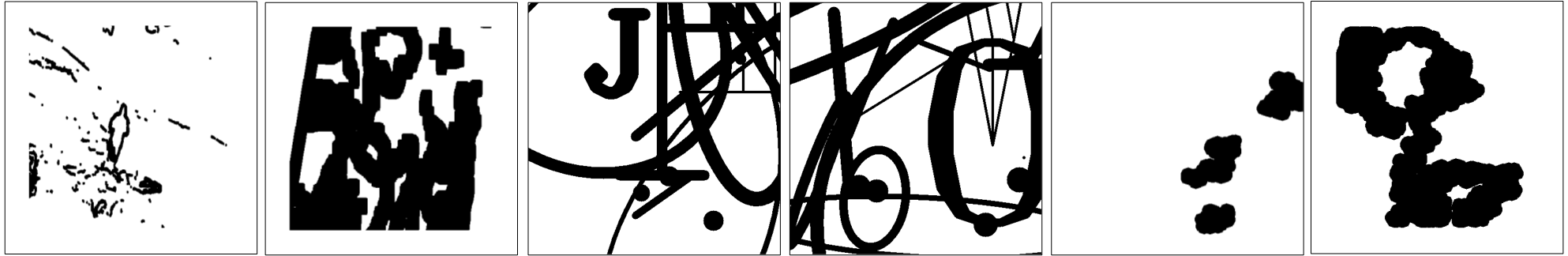}
\end{center}
\vspace{-0.3cm}
   \caption{Masks examples got by different methods. 1, 2 are from the released dataset in \cite{liu2018image}. 3,4 are created by producing kinds of random elements automatically on the canvas. 5,6 are generated using area growing crazily from a random position.}
\label{Figure 5}
\end{figure}

Different from the general generative networks where $3\times3$ or $5\times5$ convolutional filter is the most frequently applied form, deep inception models can embed larger convolutional kernels. To satisfy spatial support, dilated convolution which allows to compute each output pixel with a much larger input area while still using the same amount of parameters and computational power is adopted in \cite{iizuka2017globally}. They believe that by using dilated convolutions at lower resolutions, the model can effectively “see” a larger area of the input image when computing each output pixel than with standard convolutional layers. A larger receptive field can be obtained by adopting inception learning. And in this way, not only spatial support is satisfied, but also different views of images is obtained and  different features are learned. In order to avoid the expansion of parameters and calculations,  large convolution kernels such as $n \times n$ filter are decomposed into $n\times1$ and $1\times n$ forms. This helps save parameters, speed up calculations, and avoid over-fitting. Considering that pooling is embedded in Inception, this not only helps keep the ability of effective learning characteristics, but also avoids blurring the details and reducing the resolution caused by using pooling alone. 

Apart from parallel combination of filters, there are also cascade forms for constructing inception. By cascading convolutions, more nonlinear features are acquired. Take a two convolutions cascade combination for example, suppose the first $3\times3$ convolution + activation function approximates $f_{1}(x)=ax^{2}+bx+c$, and the second $1\times1$ convolution+activation function approximates $f_{2}(x)=mx^{2}+nx+q$, obviously, the nonlinearities of $f_{2}(f_{1}(x))$ is stronger than that of $f_{1}(x)$, the cascaded form of convolutions can simulate nonlinear features better than vanilla convolutions. Figure \ref{Figure 2} shows one kind of inception in our model.  This structure stacks convolutions ($1\times1$ , $3\times3$, $5\times5$) and pooling ($3\times3$) which are commonly used in CNNs. The dimensions of convolution and pooling are kept same and their channels are concatenated. The learned features of the examples of different kernel in the micro network are extracted for display. The visualized results are shown in Figure \ref{Figure 3}. Here, we choose the first 15 channels' outputs of one layer in the decoder stage. For a given input image shown in Figure \ref{Figure 3}(a), the feature maps of the inception are shown in Figure \ref{Figure 3}(b).
\begin{figure*}[!htbp]
\centering
\subfigure[Input]{
\begin{minipage}[t]{0.155\linewidth}
\includegraphics[width=1.0\linewidth]{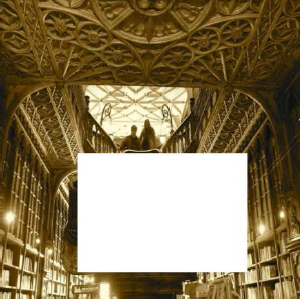}\vspace{1pt}
\includegraphics[width=1.0\linewidth]{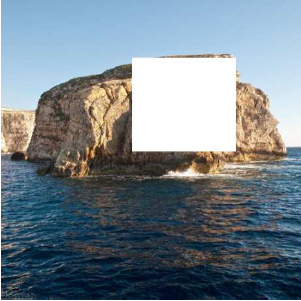}\vspace{1pt}
\includegraphics[width=1.0\linewidth]{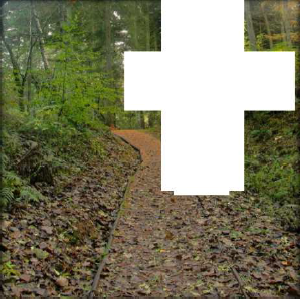}
\end{minipage}}
\subfigure[GL]{
\begin{minipage}[t]{0.155\linewidth}
\includegraphics[width=1.0\linewidth]{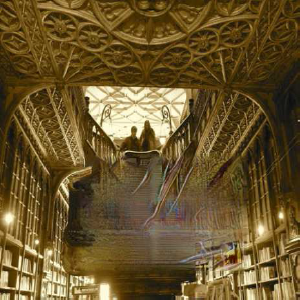}\vspace{1pt}
\includegraphics[width=1.0\linewidth]{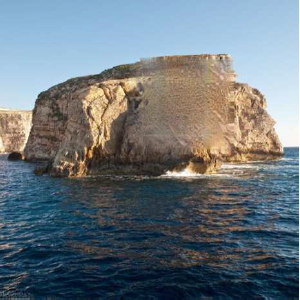}\vspace{1pt}
\includegraphics[width=1.0\linewidth]{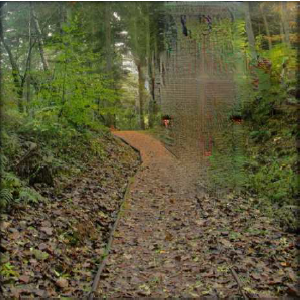}
\end{minipage}}
\subfigure[GntIpt]{
\begin{minipage}[t]{0.155\linewidth}
\includegraphics[width=1.0\linewidth]{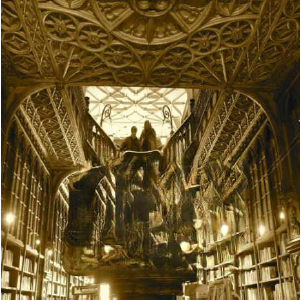}\vspace{1pt}
\includegraphics[width=1.0\linewidth]{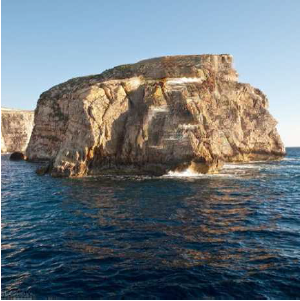}\vspace{1pt}
\includegraphics[width=1.0\linewidth]{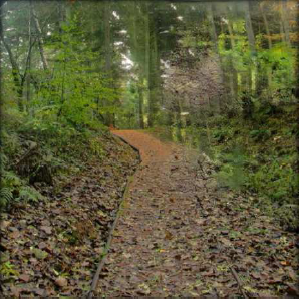}
\end{minipage}}
\subfigure[Pconv]{
\begin{minipage}[t]{0.155\linewidth}
\includegraphics[width=1.0\linewidth]{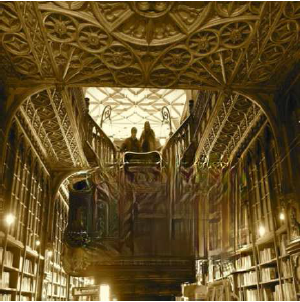}\vspace{1pt}
\includegraphics[width=1.0\linewidth]{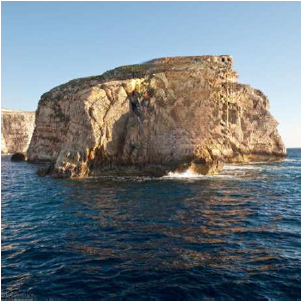}\vspace{1pt}
\includegraphics[width=1.0\linewidth]{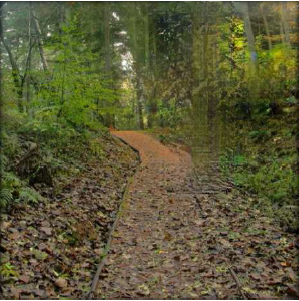}
\end{minipage}}
\subfigure[Ours]{
\begin{minipage}[t]{0.155\linewidth}
\includegraphics[width=1.0\linewidth]{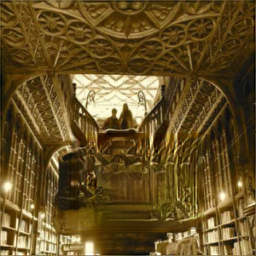}\vspace{1pt}
\includegraphics[width=1.0\linewidth]{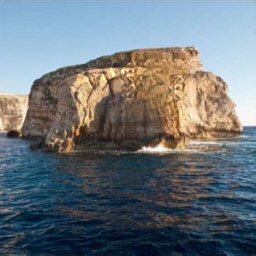}\vspace{1pt}
\includegraphics[width=1.0\linewidth]{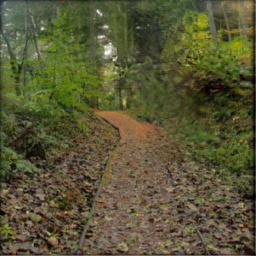}
\end{minipage}}
\subfigure[GT]{
\begin{minipage}[t]{0.155\linewidth}
\includegraphics[width=1.0\linewidth]{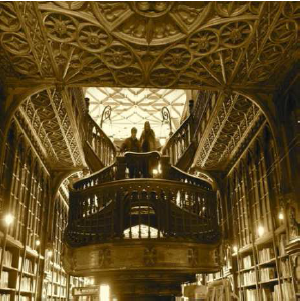}\vspace{1pt}
\includegraphics[width=1.0\linewidth]{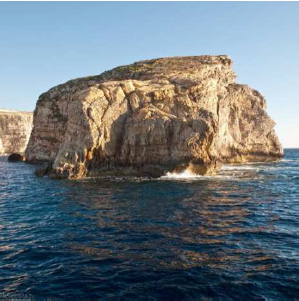}\vspace{1pt}
\includegraphics[width=1.0\linewidth]{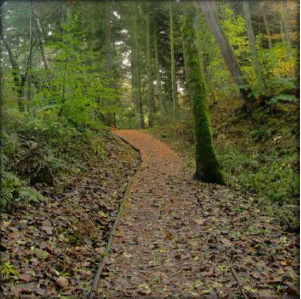}
\end{minipage}}
   \caption{Comparisons on regular masks.}
\label{Figure 6}
\end{figure*}

Such a network layer with an excellent local topology performing multiple convolution or pooling in parallel  promotes learning of multiple features. Networks constructed by such layers can better abstract image characterization. And the intuition of effective multi-scale feature representation helps bring enough cognitive understanding. Based on the aforementioned analysis, image generative networks based on deep inception learning incline to improve cognitive logicality, make higher-quality inpainting results and ameliorate the of problem of artifacts, content discrepancy, color non-consistency and discrepancy which exit in previous works due to the lack of high level context representations and non-logic cognition. 
\subsection{Partial Convolution}
\cite{liu2018image} proposed the concept of partial convolutions. It has shown outstanding performance for  images inpainting with irregular holes. The convolution is masked and the outputs are conditioned on only valid pixels. Let $W$ be the weight of the convolution filter and $b$ be the corresponding deviation. $X$ are the feature values of the current sliding window, $M$ is the corresponding binary mask.  The partial convolution at each position is expressed as:
\begin{equation}
x^{'}={\left\{
             \begin{array}{lr}
             W^T({X} \bigodot {M})\dfrac{1}{sum{M}}, sum(M)>0 &  \\
             0, otherwise &
             \end{array}
\right.}
\end{equation}
where $\bigodot$ denotes element-wise multiplication. It can be seen that the output values depend only on the unmasked region. Partial convolution has a better effect than standard convolution in correctly processing irregular masks. 
Different from image classification and object detection,  where all pixels of input image are valid, while there are invalid pixels in the holes or the masked regions for the task of image inpainting. The pixel values of the masked regions are set 0 or 255 generally. This would lead to color discrepancy, edge responses if these invalid pixels take part in convolution as reported in \cite{liu2018image}. To utilize the advantages of the recent work, we replace standard convolution with partial convolution in our model.
\subsection{Network Design} 
We follow Encoder-Decoder architecture to design the generative network. In our model, there are 16 layers totally, and 8 layers in encoder and 8 layers decoder respectively. The Encoder stage is to learn image features and it is a process of characterizing images. The Decoder stage is the process of restoring and decoding previously learned features to real images. On some issues, the information provided by the pixels around a pixel is always taken into account. This information generally consists of two categories: one is the overall environmental field information, and the other is the detailed information. The chosen form of window will bring a large uncertainty. If the selected size is too large, not only more pooling layers are required to make the environmental information appear, but also the local details are lost. On the contrary, field information is not accurate. Recent works demonstrate that U-net proposed in \cite{ronneberger2015u} has a good performance for image generative tasks. U-net uses a network structure that includes down-sampling and up-sampling. Down-sampling is used to gradually reveal the environmental information, and the up-sampling process merges the learned features which include the environmental information during down-sampling to restore more details. 
We combine the benefit of the approach in our model and the overview of the image generative architecture is shown in Figure ~\ref{Figure 4}.
\begin{figure}[!htbp]
\begin{center}
    \subfigure[]{\includegraphics[width=0.3\linewidth]{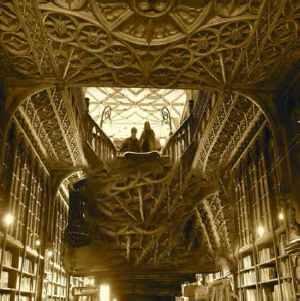}}
    \subfigure[]{\includegraphics[width=0.3\linewidth]{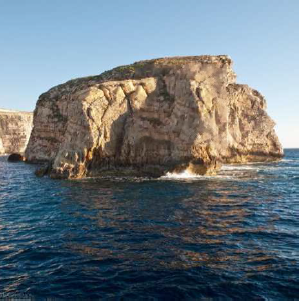}}
    \subfigure[]{\includegraphics[width=0.3\linewidth]{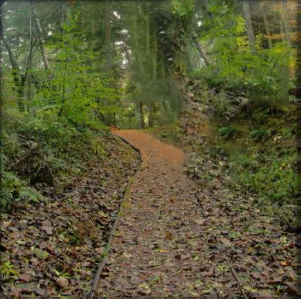}}
\end{center}
   \caption{Results by IM. a, b, c corresponds the results of the above three examples respectively.}
\label{Figure 7}
\end{figure}

In our model,  middle layers are with micro networks which are inception modules. Each convolutional layer is followed with batch normalization and activation. ReLU is for encoding layers and LeakyReLU with alpha = 0.2 is used in the decoding stage. All convolutional layers are replaced with partial convolution. A key point worth mentioning is that feature map size does not vary linearly layer by layer in the architecture. This is different from previous generative networks where feature map size changes with a factor of 2 between layers. A instance is that 4th layer and 5th layer in the encoder have the same size of $32\times32$. It is the same for bottom 4th layer and bottom 5th layer in the decoding stage. The network details are in the supplementary materials.

\subsection{Guided Objective Function}
Given a ground truth image $x$, generator $F$ produces an output $F(x)$. Let $\hat{M}$ be a binary mask corresponding to the dropped image region with a value of 0 wherever a pixel was dropped and 255 for input pixels. Generally, a normalized L1 distance is used as reconstruction loss and  is constructed on pixel-level as the follows:
\begin{equation}
  \emph{L}_{rec}{(x)}= {\left \| {\hat{M}\bigodot (x-F({(1-\hat{M})}\bigodot x ) ) }\right \|}_1
\end{equation}
\begin{figure*}[!htbp]
\centering
\subfigure[Input]{
\begin{minipage}[b]{0.19\linewidth}
\includegraphics[width=1\linewidth,height=4cm]{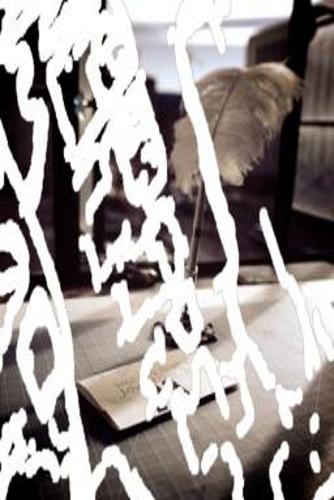}\vspace{2pt}
\includegraphics[width=1\linewidth]{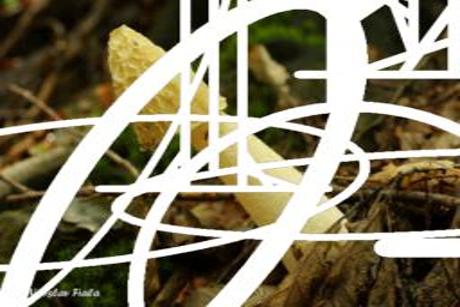}\vspace{2pt}
\includegraphics[width=1\linewidth]{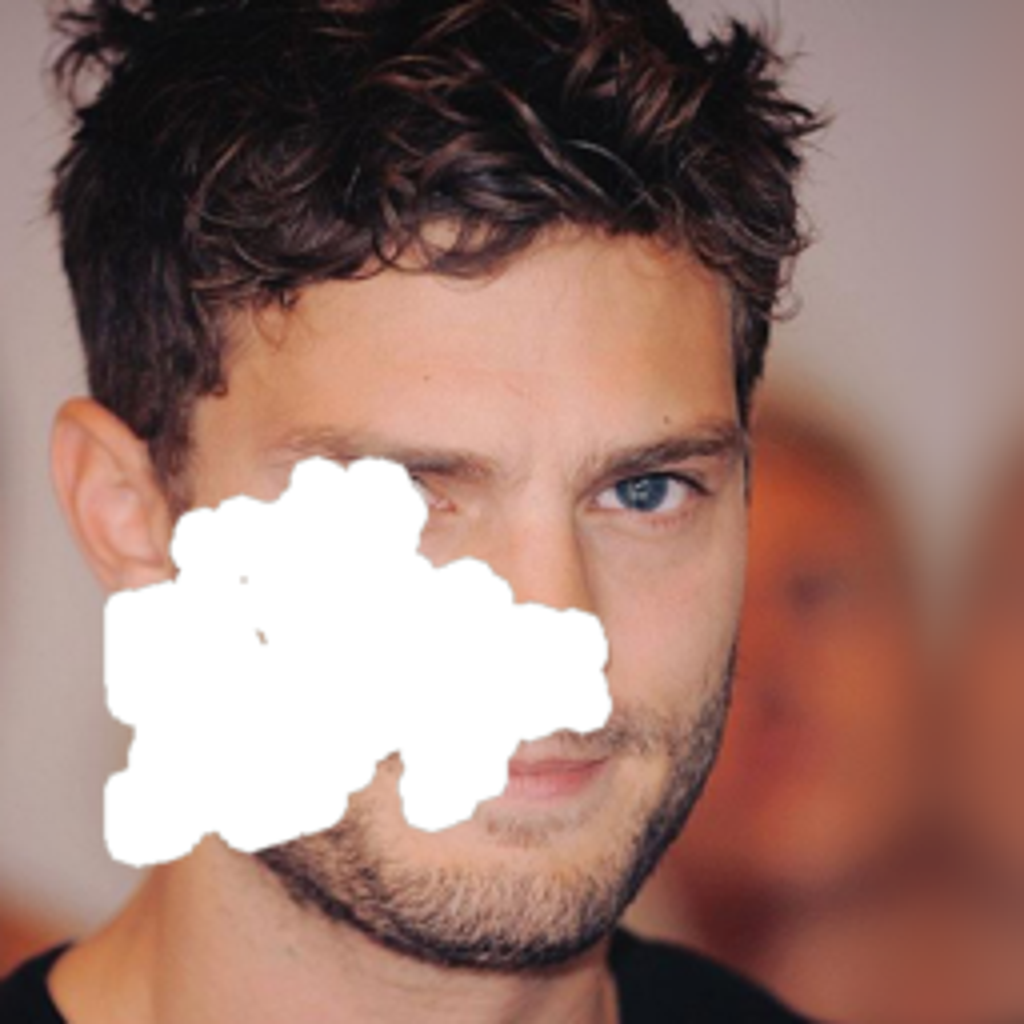}
\end{minipage}}
\subfigure[IM]{
\begin{minipage}[b]{0.19\linewidth}
\includegraphics[width=1\linewidth,height=4cm]{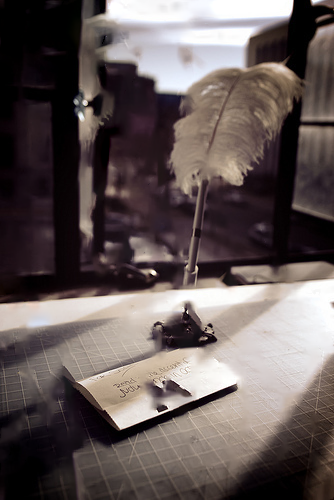}\vspace{2pt}
\includegraphics[width=1\linewidth]{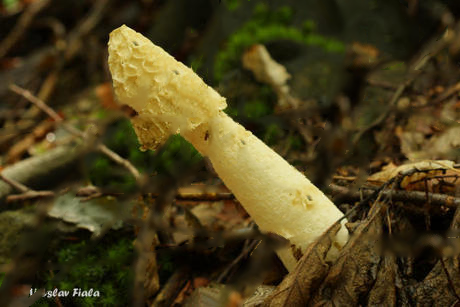}\vspace{2pt}
\includegraphics[width=1\linewidth]{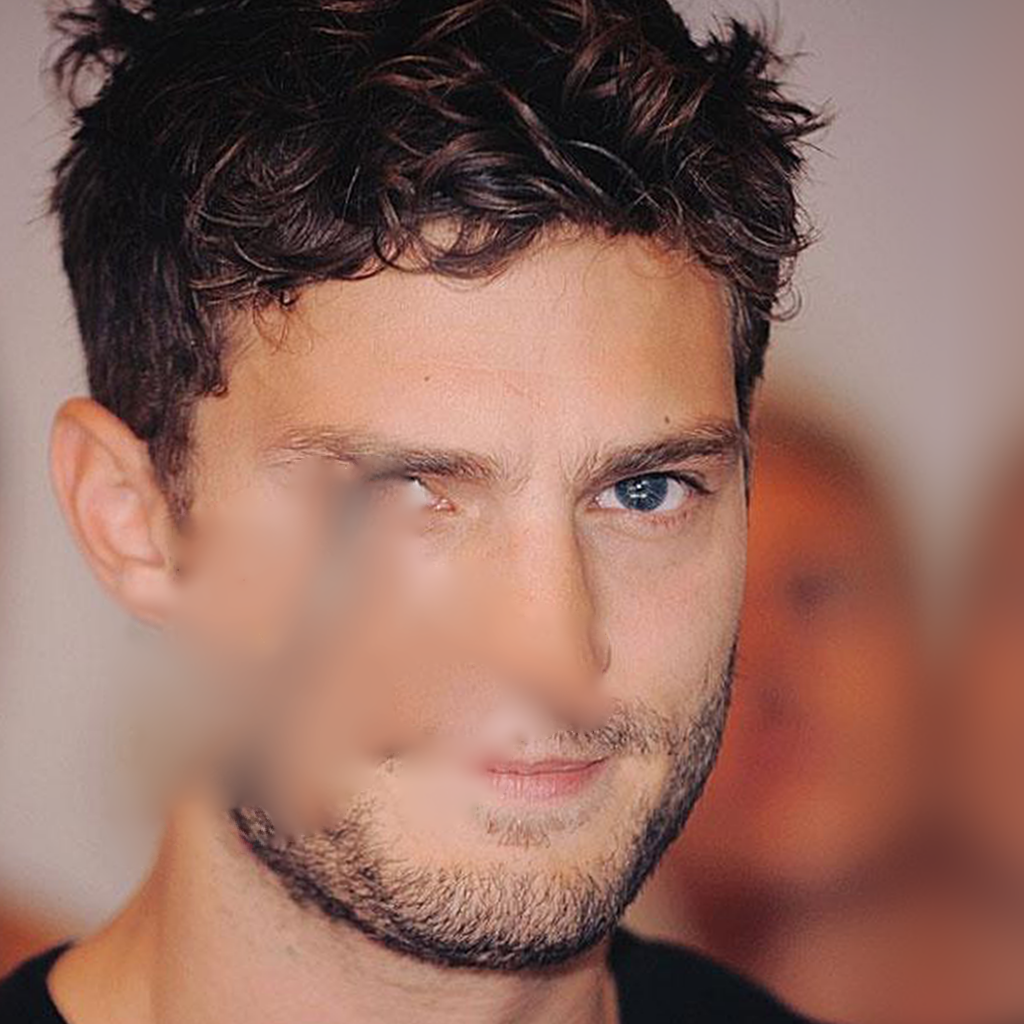}
\end{minipage}}
\subfigure[GntIpt]{
\begin{minipage}[b]{0.19\linewidth}
\includegraphics[width=1\linewidth,height=4cm]{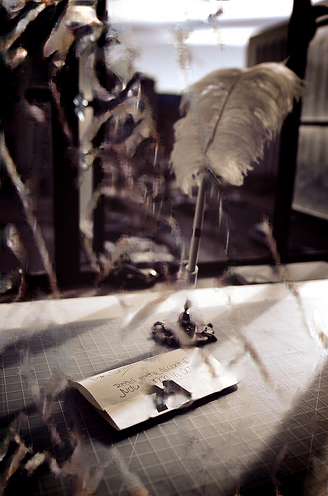}\vspace{2pt}
\includegraphics[width=1\linewidth]{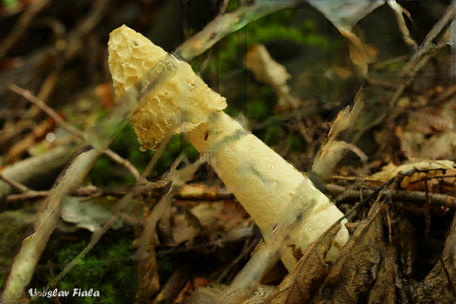}\vspace{2pt}
\includegraphics[width=1\linewidth]{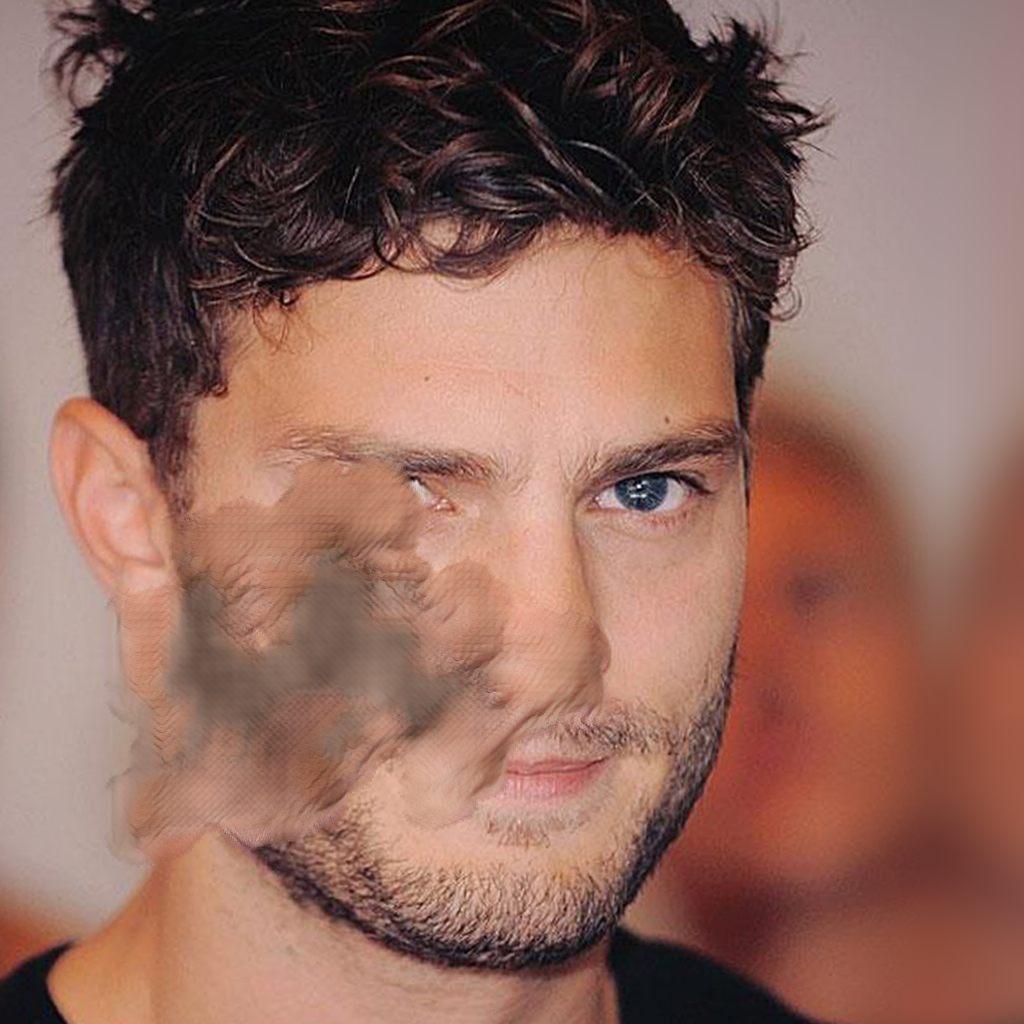}
\end{minipage}}
\subfigure[Ours]{
\begin{minipage}[b]{0.19\linewidth}
\includegraphics[width=1\linewidth,height=4cm]{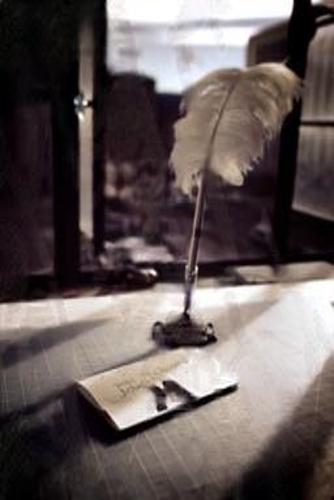}\vspace{2pt}
\includegraphics[width=1\linewidth]{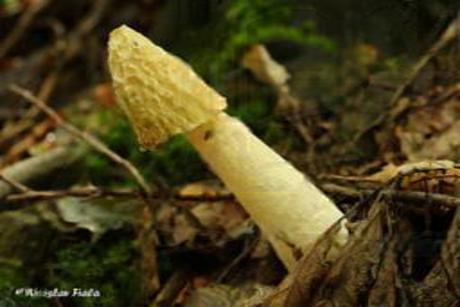}\vspace{2pt}
\includegraphics[width=1\linewidth]{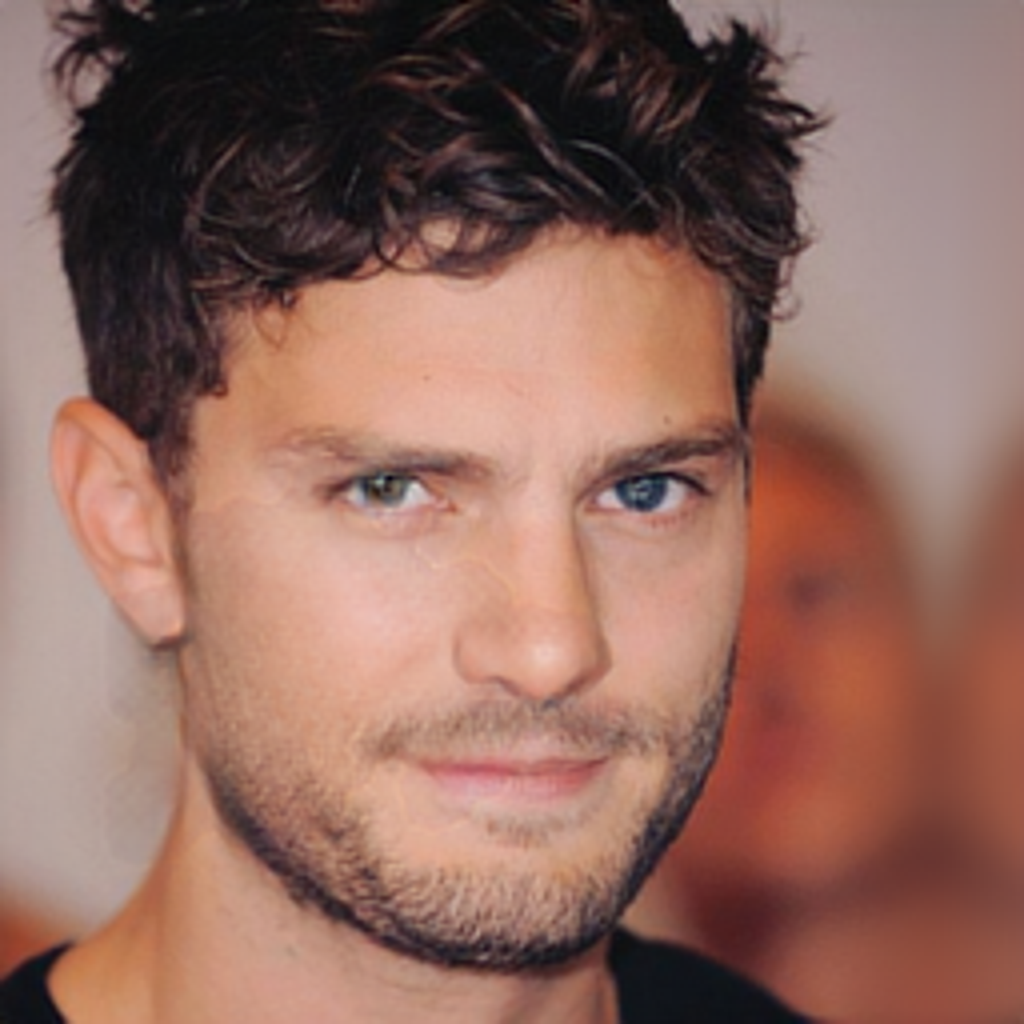}
\end{minipage}}
\subfigure[GT]{
\begin{minipage}[b]{0.19\linewidth}
\includegraphics[width=1\linewidth,height=4cm]{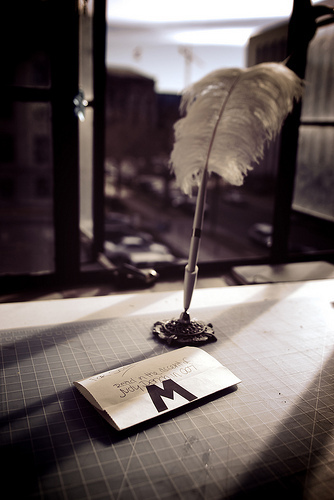}\vspace{2pt}
\includegraphics[width=1\linewidth]{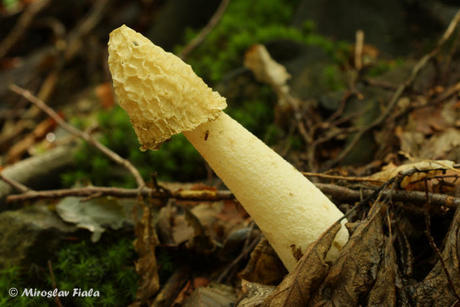}\vspace{2pt}
\includegraphics[width=1\linewidth]{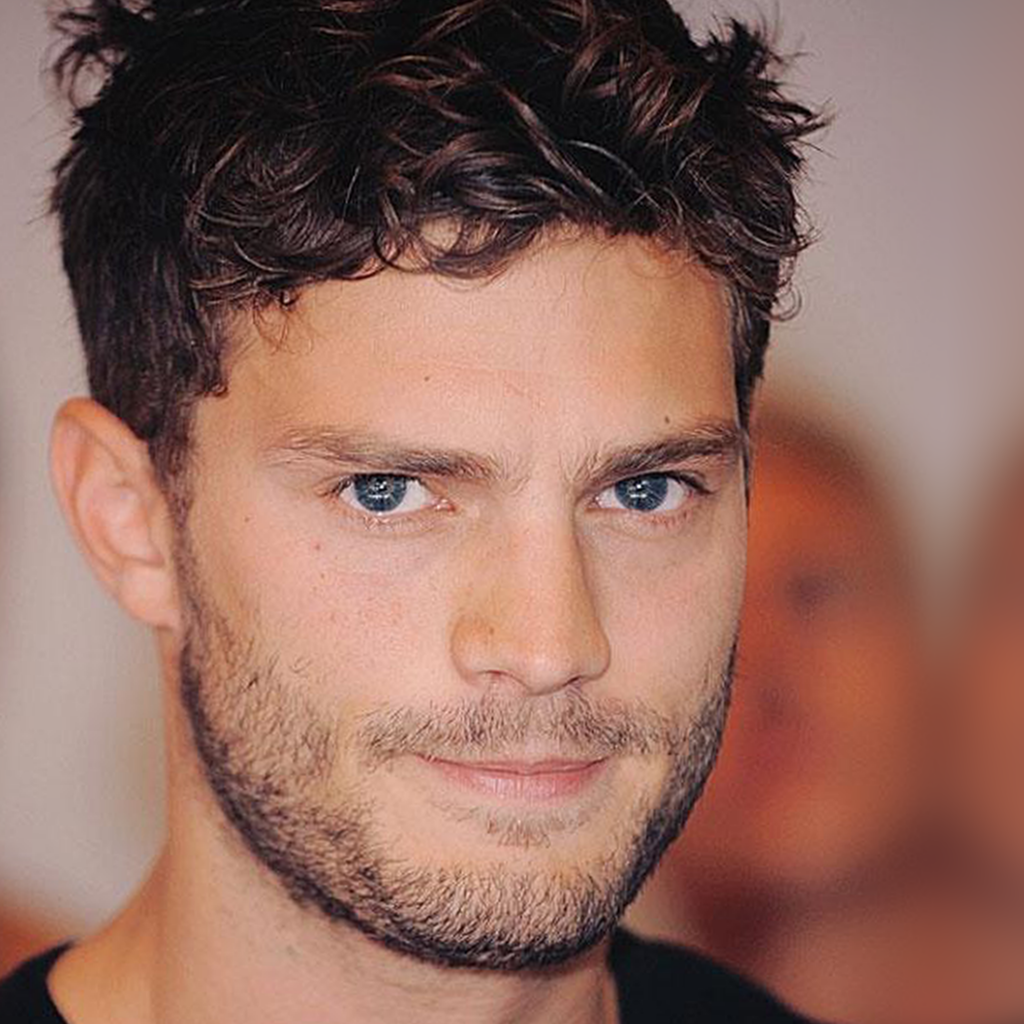}
\end{minipage}}
\caption{Comparisons on irregular masks.  }
\label{Figure 8}
\end{figure*}

Perceptual loss is first proposed in \cite{johnson2016perceptual} for real-time style transfer and super-resolution reconstruction. It is based on pre-trained networks and defined as:
\begin{equation}
\ell_{perc}^{\phi,j}(\hat{y} ,y)= \frac{1}{C_{j}H_{j}W_{j}}{\lVert {\phi}_{j}{\hat{y}} - {\phi}_{j}{y} \rVert}_1
\end{equation}
where ${\phi}_{j}$ is the feature map of the $nth$ selected layer of the pre-trained model which is usually VGG \cite{simonyan2014very}.
Perceptual loss is first applied to image inpainting in \cite{yang2017high}. Results indicate that it helps infer more accurate reconstruction of the hole content.

Style loss is widely used in image-to-image translation tasks such as real-time style transfer. It is also adopted in this paper. By computing gram matrix on each feature map, style loss takes the following form:
\begin{equation}
\ell_{style}^{\phi,j}(\hat{y} ,y)= {\lVert G_{j}^{\phi}{\hat{y}} - G_{j}^{\phi}{y} \rVert}_1
\end{equation}
where $G_{j}^{\phi}$ is the computed gram matrix.

The above tow loss functions based on high-level features extracted from pre-trained networks are combined into the final guided objective function for generating high-quality images. Finally,  the objective function takes the form:  a low-level loss based on L1 computation is for pixel distance and  a pre-trained VGG network is adopted to extract high-level feature maps for calculating the perceptual and style differences. 

\subsection{Random Mask Dataset}  
Masks play an important role in arbitrary completion and free-style inpainting. A key point is the generation of random mask images, since the deep learning-based method requires mask samples in advance other than unlabeled real images.  A latest work demonstrates that basing on the results of the occlusion/non-occlusion mask image estimation method between two consecutive frames of the video can generate better masks of arbitrary shapes and stripes \cite{sundaram2010dense}. Based on this method, a random mask dataset is created in \cite{liu2018image}. Two algorithms for automatically generating random  masks are introduced in this paper. First, pictures with multiple shapes including rectangles, circles, ellipses, and strings are randomly draw. They are randomly generated with random size, rotation, and position. Another approach is that a point is first selected randomly, then a damaged region around the point is expanded wildly. Morphology process of dilation is applied to produce the final binary mask. A mask dataset for free-style inpainting is finally created. Samples of the dataset are shown in Figure ~\ref{Figure 5}. Note that holes are black and represented with 0 in the experiments. We inverse them for demonstrating comparisons. The created random mask dataset and the code will be released on github. 

\section{Experiments and Results} 
We perform experiments on three datasets, ImageNet \cite{russakovsky2015imagenet}, Places2 dataset \cite{zhou2018places}, CelebA-HQ \cite{karras2017progressive}. The experiments are conducted with two NVDIA Geforce GTX-1080 Ti GPUs. The adopted deep learning platform is Pytorch. And we use Adam \cite{kingma2014adam} for optimization with a batch size of 16. Image samples are resized to $256\times256$ before fed to the network.
\begin{figure*}[!htbp]
\centering
\subfigure[Input]{
\begin{minipage}[b]{0.19\linewidth}
\includegraphics[width=1\linewidth]{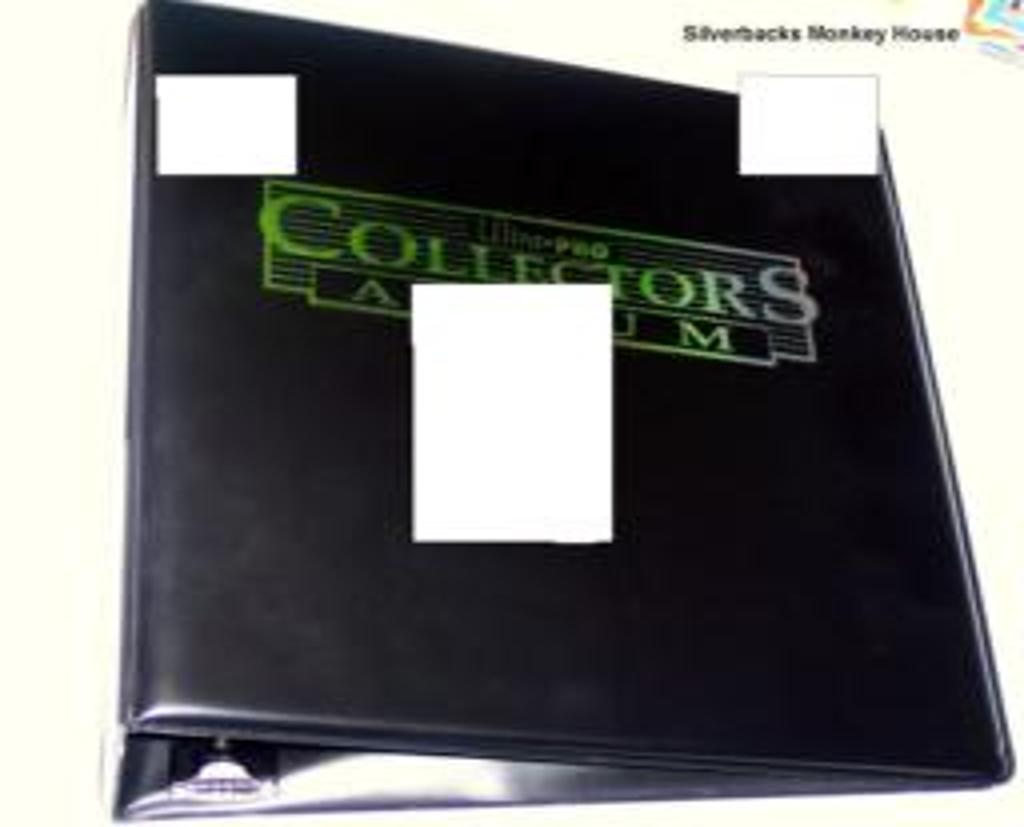}\vspace{2pt}
\includegraphics[width=1\linewidth,height=3.5cm]{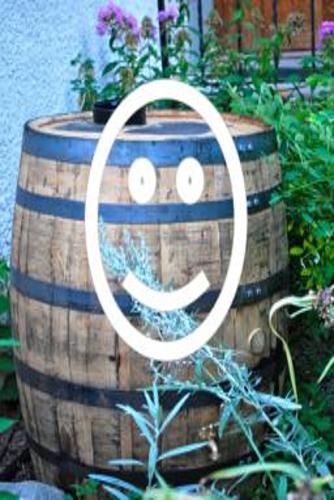}\vspace{2pt}
\includegraphics[width=1\linewidth]{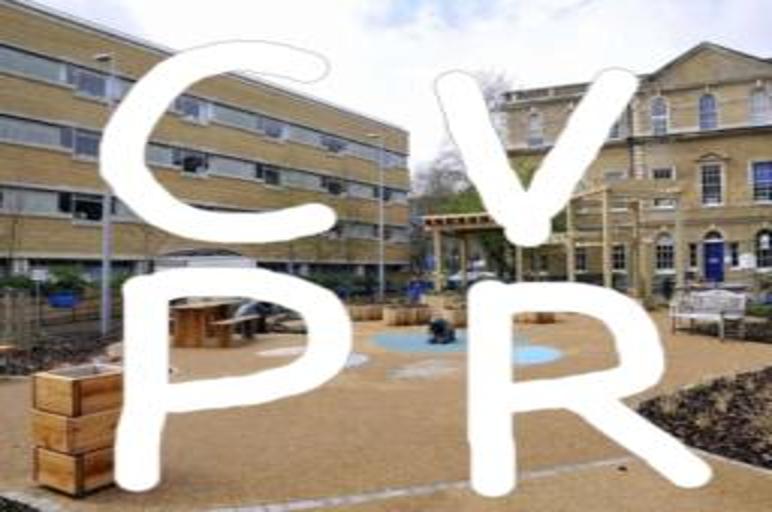}
\end{minipage}}
\subfigure[GL]{
\begin{minipage}[b]{0.19\linewidth}
\includegraphics[width=1\linewidth]{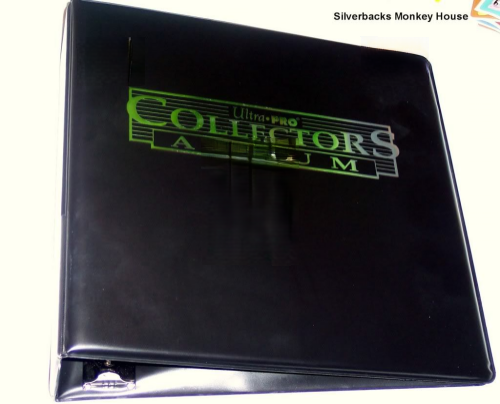}\vspace{2pt}
\includegraphics[width=1\linewidth,height=3.5cm]{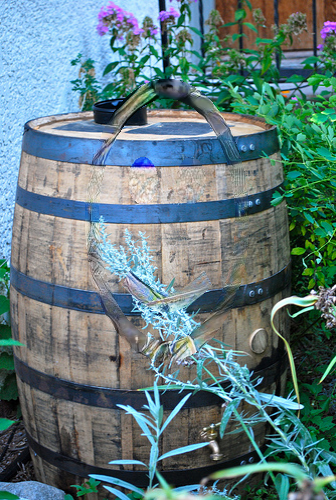}\vspace{2pt}
\includegraphics[width=1\linewidth]{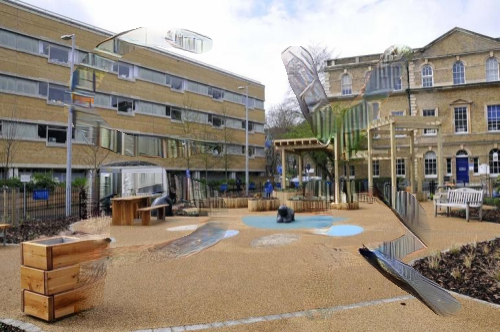}
\end{minipage}}
\subfigure[GntIpt]{
\begin{minipage}[b]{0.19\linewidth}
\includegraphics[width=1\linewidth]{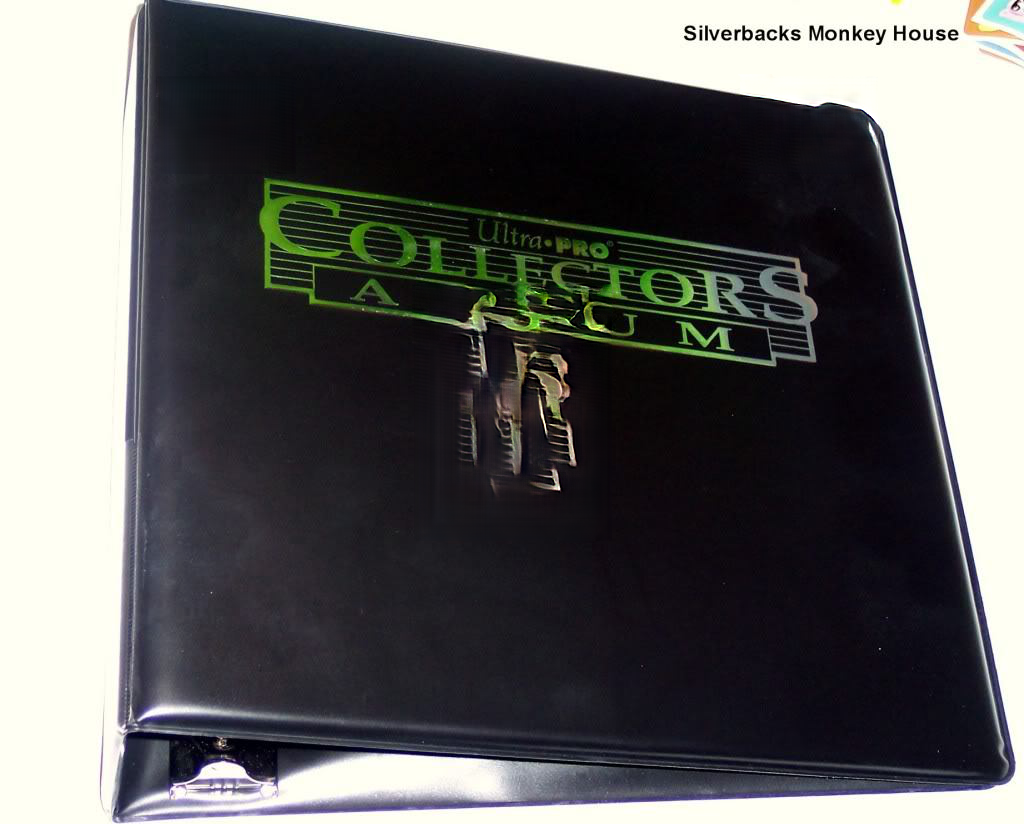}\vspace{2pt}
\includegraphics[width=1\linewidth,height=3.5cm]{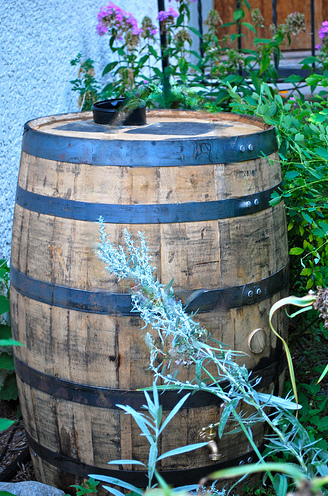}\vspace{2pt}
\includegraphics[width=1\linewidth]{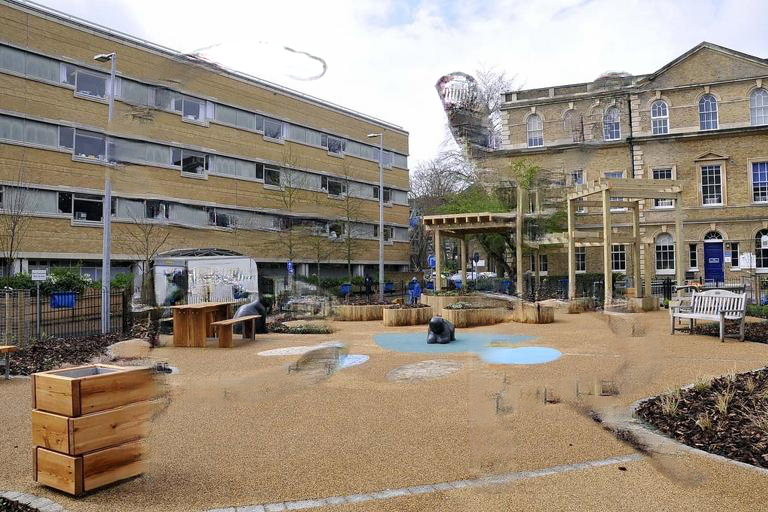}
\end{minipage}}
\subfigure[Ours]{
\begin{minipage}[b]{0.19\linewidth}
\includegraphics[width=1\linewidth]{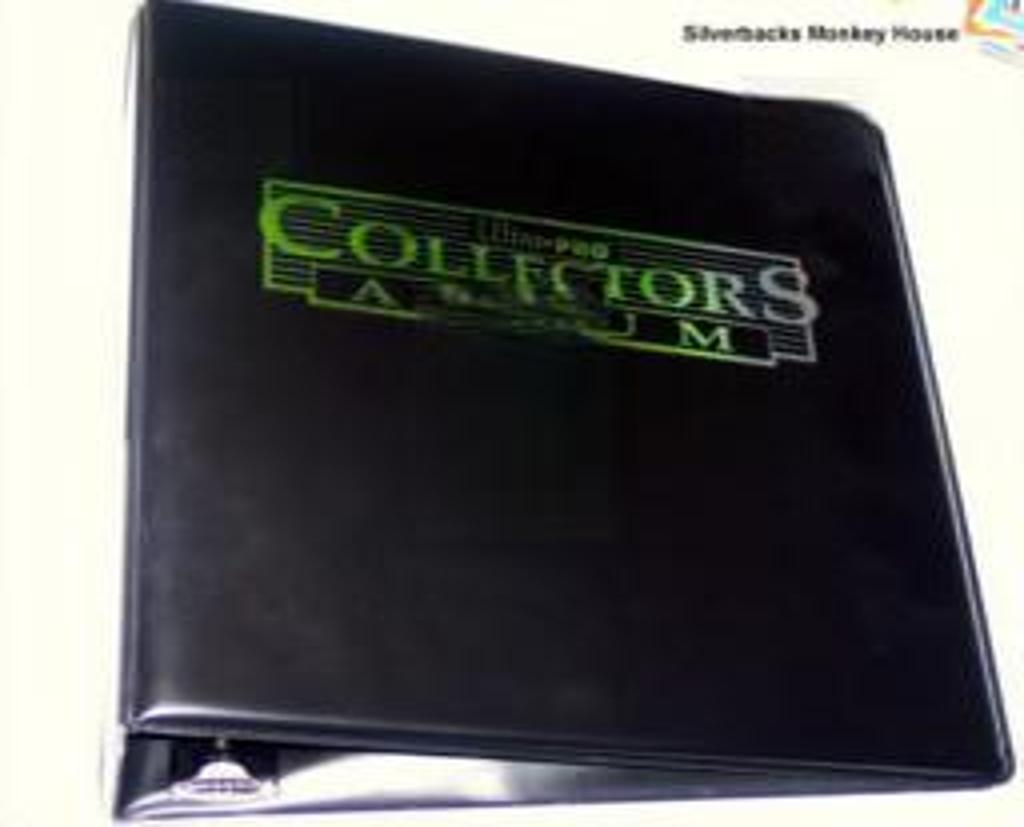}\vspace{2pt}
\includegraphics[width=1\linewidth,height=3.5cm]{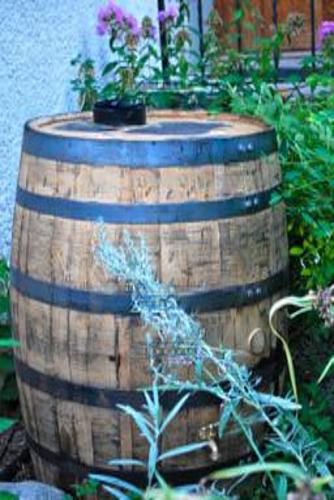}\vspace{2pt}
\includegraphics[width=1\linewidth]{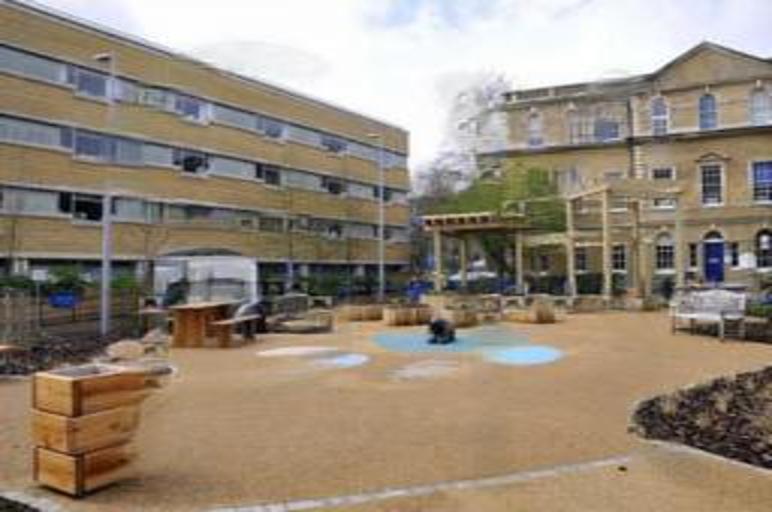}
\end{minipage}}
\subfigure[GT]{
\begin{minipage}[b]{0.19\linewidth}
\includegraphics[width=1\linewidth]{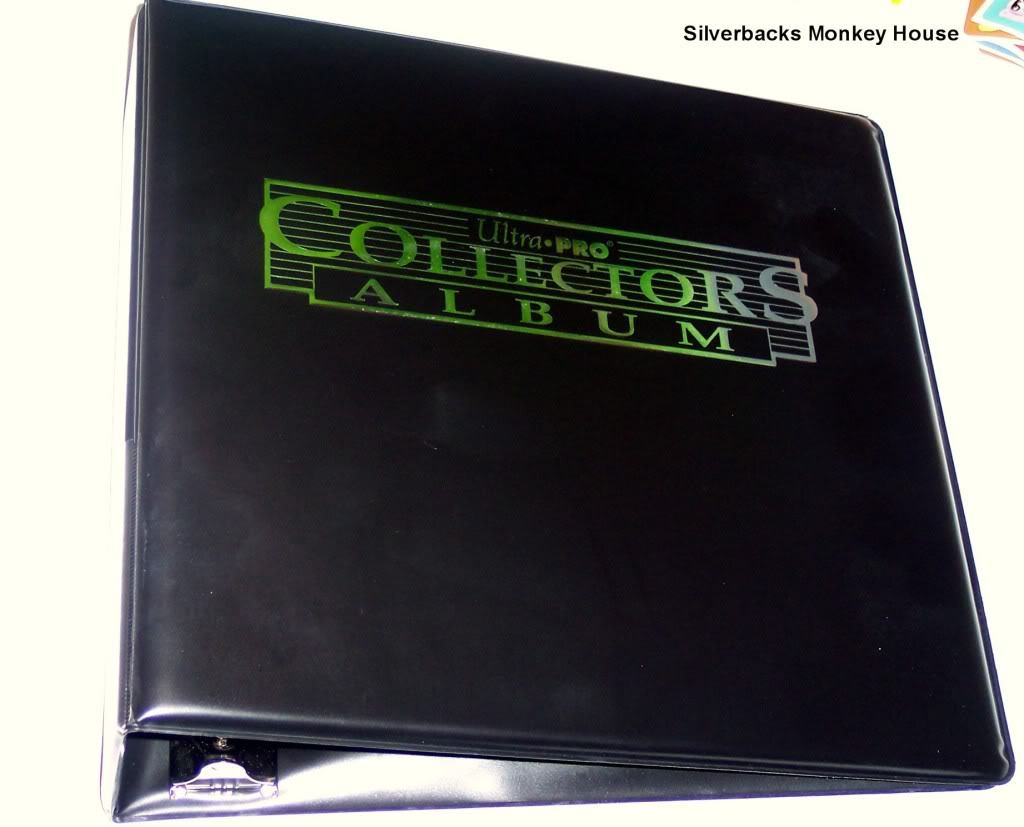}\vspace{2pt}
\includegraphics[width=1\linewidth,height=3.5cm]{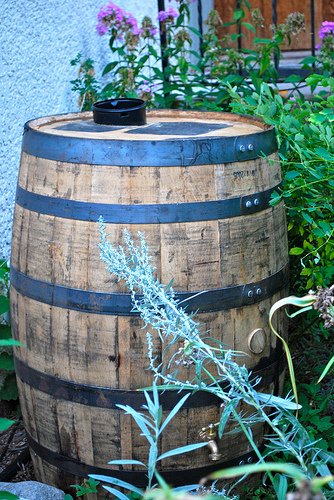}\vspace{2pt}
\includegraphics[width=1\linewidth]{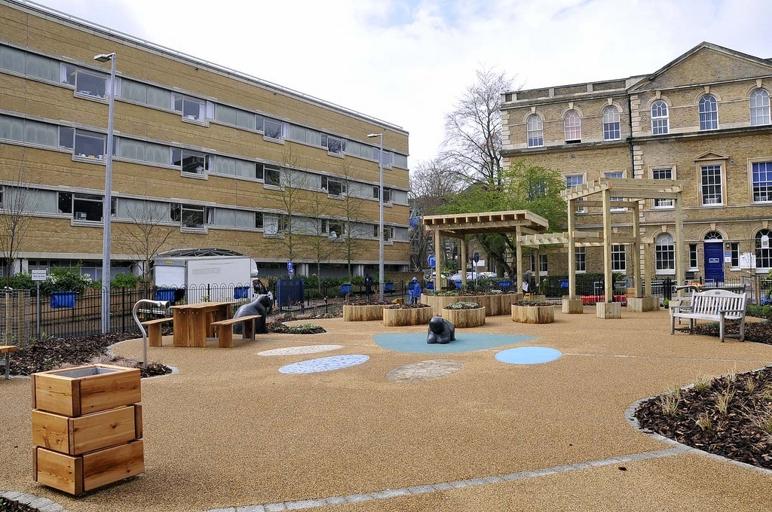}
\end{minipage}}
\label{Figure 9}
\caption{Examples of free-style inpainting. The mask are arbitrary.}
\end{figure*}

We carry out quantitative comparisons with state-of-the art algorithms, IM \cite{darabi2012image}, GL \cite{iizuka2017globally}, GntIpt \cite{yu2018generative}, Pconv \cite{liu2018image}. The implementations of all these methods were based on their released source codes, pre-trained models or results in their papers.
\subsection{Regular Regions Inpainting}   
Considering that there is no released codes and pre-trained models for Pconv, we directly use the paper results in \cite{liu2018image} for this comparison. Comparisons of different methods are shown in Figure ~\ref{Figure 6}. It can be seen that GL and GntIpt fail to achieve plausible results even through post-processing or refinement network. The two create distorted  structures inconsistent with surrounding areas because of insufficient understanding of the image characteristics and ineffectiveness of convolutional neural networks in modeling long-term correlations between contextual information and the hole regions. There are many checkerboard artifacts in Pconv while ours is less compared with it.
\begin{figure}[!htbp]
\begin{center}
\subfigure{\includegraphics[width=0.4\linewidth]{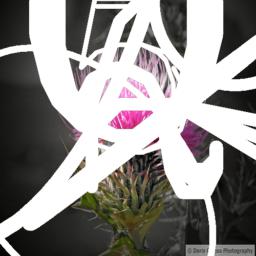}}
\subfigure{\includegraphics[width=0.4\linewidth]{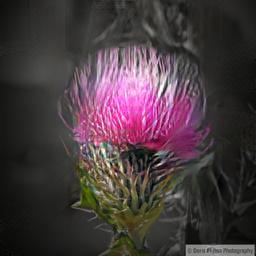}}
\subfigure{\includegraphics[width=0.4\linewidth]{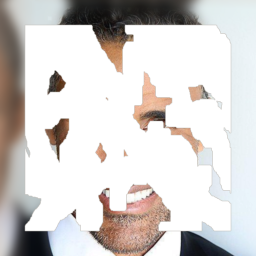}}
\subfigure{\includegraphics[width=0.4\linewidth]{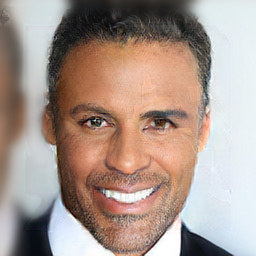}}
\end{center}
   \caption{Additional image completion results by our approach using randomly generated masks.}
\label{Figure 10}
\end{figure}
\subsection{Irregular Regions Inpainting} 
For the comparison, we used the released codes in \cite{darabi2012image} for IM and pre-trained models in \cite{yu2018generative}. From the results, we can see that the results of IM are receivable for  filling narrow or small holes. However, its result is blurry for filling large holes as shown in the last row of Figure ~\ref{Figure 8}. The completed contents are not consistent with the scene and the copied patches from somewhere else are disharmonious with the surroundings due to that it is unable to generate novel objects in the image as shown in Figure ~\ref{Figure 7}.  
\subsection{Free-style Inpainting}
The mask images used here are got manually. Some figures are draw by hand on canvas. Owing to insufficient cognitive understanding, GL and GntIpt create distorted structures and artifacts as shown in Figure ~\ref{Figure 9}. Although GL works well in the first row and the second result of GntIpt is receivable, they fail in other instances. The results do not comply with our human vision cognition. Note that different from previous work where the mask samples in test have the same probability distribution with that in train because they are generated using the same way, the distribution and peculiarity of the used mask samples in free-style inpainting are previously unknown and the masks can be created by our freewill. Results demonstrate that our method generates  much more natural image completion for kinds of mask samples. More results are in Figure \ref{Figure 10}.

\section{Discussion}
In this paper, deep inception learning is adopted for cognitive inpainting. Combined with the benefits of some recent approaches, a state-of-the-art generative network is designed. Furthermore, two approaches for generating random mask samples are introduced. We valid our methods on three benchmark datasets. Experiments show that superior inpainting results are obtained and our method is robust for diverse masks which are vital for free-style inpainting. As image inpainting has commonality with super-resolution tasks, we plan to extend our built model to image super-resolution reconstruction.

{\small
\bibliographystyle{ieee}
\bibliography{egpaper_for_review_arxiv.bbl}
}
\end{document}